\documentclass[
anonymous
]{ceurart}

\usepackage{listings}
\usepackage{amsmath}
\usepackage{tabularx}
\usepackage{xcolor} 
\usepackage{booktabs} 

\begin{document}

\copyrightyear{2025}
\copyrightclause{Copyright for this paper by its authors.
  Use permitted under Creative Commons License Attribution 4.0
  International (CC BY 4.0).}

\conference{MAI-XAI'25: Workshop on Multimodal, Affective and Interactive Explainable AI
October 25–26, 2025, Bologna, Italy}

\title{Adaptive XAI in High Stakes Environments: Modeling Swift Trust with Multimodal Feedback in Human AI Teams }


\author[1]{Nishani Fernando}[%
email=nlfernando11@gmail.com]
\cormark[1]
\address[1]{Deakin University,
  Geelong, Victoria, Australia}

\author[1]{Bahareh Nakisa}[%
email=bahar.nakisa@deakin.edu.au]
\cormark[1]

\author[1]{Adnan Ahmad}[%
email=adnan.a@deakin.edu.au]
\cormark[1]

\author[2]{Mohammad Naim Rastgoo}[%
email=naim.rastgoo@monash.edu]
\address[2]{Monash University, Melbourne, Victoria, Australia}
\cormark[1]


\begin{abstract}
 Effective human-AI teaming heavily depends on swift trust, particularly in high-stakes scenarios such as emergency response, where timely and accurate decision-making is critical. In these time-sensitive and cognitively demanding settings, adaptive explainability is essential for fostering trust between human operators and AI systems. However, existing explainable AI (XAI) approaches typically offer uniform explanations and rely heavily on explicit feedback mechanisms, which are often impractical in such high-pressure scenarios. To address this gap, we propose a conceptual framework for adaptive XAI that operates non-intrusively by responding to users’ real-time cognitive and emotional states through implicit feedback, thereby enhancing swift trust in high-stakes environments. The proposed adaptive explainability trust framework (AXTF) leverages physiological and behavioral signals, such as EEG, ECG, and eye tracking, to infer user states and support explanation adaptation. At its core is a multi-objective, personalized trust estimation model that maps workload, stress, and emotion to dynamic trust estimates. These estimates guide the modulation of explanation features enabling responsive and personalized support that promotes swift trust in human-AI collaboration.
 This conceptual framework establishes a foundation for developing adaptive, non-intrusive XAI systems tailored to the rigorous demands of high-pressure, time-sensitive environments.
  
\end{abstract}



\maketitle

\section{Introduction}

In high stakes domains such as emergency response \cite{lim2021cognitive} and military operations \cite{Dehais2019MonitoringPM}, human AI teams are often formed on the fly and operate under extreme time pressure, high cognitive workload, and rapidly evolving situational demands. These environments are characterized by rapid decision-making, elevated emotional intensity, and limited opportunities for explicit communication or coordination. Failures in such contexts can lead to significant safety, ethical, or operational consequences \cite{rodriguez2023}. As a result, effective human-AI teaming in such scenarios hinges on the development of swift trust and the ability to support human operators through adaptive, context-sensitive system behavior. Swift trust, originally introduced in the context of temporary human teams \cite{Meyerson1996}, describes a form of trust that arises rapidly out of necessity, without the benefit of prolonged interaction or prior history. In high-stakes environments, humans are often compelled to place immediate trust in AI systems simply because there is no time to build it gradually. However, this initial trust is fragile and often vulnerable to performance errors, lack of transparency and high cognitive load \cite{Hancock2011}. Sustaining trust in such conditions demands that AI systems must be capable of communicating effectively and adapting responsively to the human’s evolving cognitive and emotional state.

Explainability has emerged as a central mechanism for cultivating trust in AI, enabling humans to understand and anticipate AI behavior \cite{Shin2021}. However, existing explainable AI (XAI) approaches \cite{Ali2023} are often static and uniform, providing generic explanations that overlook situational awareness and fail to adapt to the dynamic nature of the environment, which directly influences the user’s real-time cognitive and emotional state.
Furthermore, these approaches typically rely on explicit human feedback, such as verbal queries or stated preferences, which are often impractical in high-pressure, cognitively demanding environments where users are overloaded and time is constrained. Therefore, more advanced explainable systems are essential for high-stakes environments. Such systems must be capable of rapid adaptation not only to the human operator’s state but also to contextual variables like task urgency, system reliability, and environmental uncertainty.


To overcome these gaps, incorporating implicit human feedback is essential for advancing explainable AI (XAI) systems, especially in high-stakes environments. Non-invasive technologies, such as wearable sensors \cite{Akash2018}, provide a promising means of capturing physiological and behavioral signals that reflect a user’s internal state.  These signals may include Electroencephalography (EEG) \cite{Akash2018, Choo2022}, Electrocardiography (ECG) \cite{AhmadAdnan2025}, and eye tracking \cite{Hu2024, Hulle2024}, serving as real-time proxies for trust, cognitive workload, and emotional state. However, effective explanation adaptation must also account for AI system performance and situational context, which collectively shape human-AI trust dynamics. A comprehensive, adaptive XAI framework must therefore integrate these diverse signals to provide personalized, context-aware support.


This work presents adaptive explainability trust framework (AXTF), a conceptual framework designed to advance human-AI teaming in high-stakes, time-sensitive domains by enabling adaptive, non-intrusive explainability driven by multi-objective trust estimation model. It outlines a foundational approach that combines implicit human feedback, AI performance metrics, and situational awareness to infer the evolving trust state of the user. At the core of this framework is a personalized trust inference model that integrates the user’s cognitive and emotional state along with situational awareness to infer dynamic trust levels. These estimates guide the adaptive modulation of explanation features such as timing, granularity, content, and presentation mode, enabling dynamic, context-aware explanation strategies that foster swift trust. This conceptual approach lays the foundation for future research and the practical development of trust-sensitive, non-intrusive XAI systems tailored to the demands of time-critical, high-pressure domains.

Unlike task specific or opaque AI models \cite{Soni2021}, our framework is generalizable across high stakes domains, interpretable by design, and adaptable in real time. It supports collaboration by recognizing the cognitive and affective constraints of human operators and responding accordingly closing the loop between trust inference, explanation adaptation, and mission performance. In the following sections, we review prior work, detail our conceptual model, and outline its application to real-world high-pressure settings such as emergency response.

\section{Background and Related Work} \label{sec:relatedwork}

This section introduces key foundations for our proposed framework by synthesizing prior work across four core areas. First, we examine the role of implicit feedback in high-stakes human-AI collaboration, emphasizing the need for implicit, real-time cues such as physiological and behavioral signals to support decision-making under stress and cognitive load. Next, we explore the construct of swift trust, outlining its determinants, reliability, predictability, competence, transparency, and adaptability and their sensitivity to fluctuating user states. We then review the limitations of adaptive explainability, highlighting gaps in existing XAI systems that fail to adjust explanations to changing user conditions. Finally, we discuss user-centric and affect-aware XAI, emphasizing emerging evidence for modeling trust through real-time physiological inference, and outlining the need for integrated models that dynamically adapt explanation features to maintain trust and cognitive efficiency in time-sensitive contexts. This background frames the motivation and design of our proposed conceptual model.

\subsection{Human AI Collaboration and the Role of Implicit Feedback }
In high stakes domains, human AI collaboration is often task critical, requiring both agents to operate in close coordination under intense cognitive and temporal pressure. The effectiveness of this teaming rests heavily on the human operator’s ability to trust the AI system to understand its role, predict its behavior, and rely on its outputs in moments of uncertainty. Studies across domains such as emergency response and autonomous operations show that trust in AI systems enhances decision making efficiency, reduces cognitive burden, and improves overall team performance \cite{Hancock2011, Milivojevic2024,Paleja2021}. However, traditional models of trust formation often assume explicit communication between humans and AI agents, such as requests for clarification, preference adjustment, or corrective feedback. In practice, such explicit feedback is limited or infeasible in high stakes situations \cite{Endsley2023}. Human operators are typically focused on the task at hand, operating under cognitive overload, and have minimal capacity to verbally assess or tune their interaction with the AI system. This constraint necessitates an alternative trust support mechanism, one that can function implicitly, adaptively, and in real time. 

Implicit feedback, measured through physiological and behavioral signals, offers a promising foundation for adaptive support in human-machine teams (HMTs). Unlike explicit feedback, implicit indicators are passively observable and continuous, providing a non-intrusive method for assessing the user's cognitive and emotional state. A growing body of research supports the viability of using such signals for trust estimation. For instance, EEG has been shown to correlate with cognitive load and attention \cite{rodriguez2023}. Similarly, ECG and galvanic skin response (GSR) are reliable indicators of physiological arousal and stress, which have been linked to trust erosion under pressure \cite{shaffer2017, Akash2018, green2025, Aygun2020, Rastgoo2024, Nakisa2018}. Moreover, gaze patterns, facial expressions, and vocal features reflect emotional valence and engagement, serving as real-time proxies for user affect and trust \cite{Khalid2016}. These findings suggest that trust-relevant mental states can be inferred in real time from sensor data, enabling AI systems to detect when a user is confused, overloaded, disengaged, or stressed without requiring explicit articulation.

    
    
    


By grounding our model in these implicit signals, we enable AI systems to dynamically assess the operator’s cognitive and emotional state and adapt their behavior accordingly. In the context of XAI, this means tailoring explanations to be more concise, timely, or expressive depending on the inferred user state. For example, a spike in physiological arousal following an AI action may indicate confusion or concern, prompting the system to proactively issue a clarifying explanation. Similarly, indicators of high trust and low load might invite more detailed, exploratory explanations to support learning or calibration. In this way, implicit feedback enables real time, user sensitive adaptation, forming a critical bridge between human trust dynamics and machine explainability. It allows the AI system to act as an responsive teammate not just explaining what it does, but choosing when and how to explain based on the operator’s needs. This perspective forms the foundation for the next section, where we examine the elements of swift trust, and how they intersect with user state and explainability in high stake team settings. 

\subsection{Swift Trust and Its Determinants }

In high stakes human AI collaboration, trust must be formed quickly often in the absence of prolonged interaction or past performance history. This phenomenon, referred to as swift trust \cite{Meyerson1996}, is essential for enabling rapid coordination in dynamic environments such as disaster response or critical medical care. Unlike traditional trust, which emerges gradually through relationship building, swift trust is assumed provisionally based on contextual cues like system role, professionalism, and perceived competence. However, swift trust is inherently fragile. It can erode rapidly when system behavior is unclear, inconsistent, or perceived as unreliable. Maintaining and calibrating this trust is a non-trivial challenge especially under conditions of high workload and emotional strain, where human perception of system behavior becomes volatile. A breakdown in trust can lead to over reliance (complacency) or under reliance (disuse), both of which are detrimental to team performance. 




A large body of empirical work (e.g.{\cite{Milivojevic2024, kohn2021, Paleja2021}) confirms that trust in AI systems is closely linked to perceived performance, system reliability and predictability, as well as the user’s workload, stress, and emotional state. Hancock et al. \cite{Hancock2011} found that performance was the strongest predictor of trust, with workload and environmental risk also contributing significantly. More recent work shows that stress and cognitive overload reduce perceived trust, while positive emotional states such as engagement promote confidence and trust \cite{rodriguez2023, Sadrfaridpour2014, Paleja2021}. These variables are not only correlational but causal, influencing how operators interpret and respond to AI recommendations in real time. For example, when workload is high and system behavior is unclear, trust may drop even if the AI is performing correctly. Conversely, under calm conditions with transparent AI behavior, trust can remain stable even after minor failures. Trust, therefore, operates as a feedback variable, modulated by both system performance and the human’s cognitive and emotional state. Hoff and Bashir’s \cite{Hoff2015} three-level trust model highlights that initial trust or disposition to trust is shaped by individual traits, prior experiences, and cultural factors, serving as a baseline for interaction with automation systems. While these dispositional influences are important, our work focuses on the dynamic adaptation of trust during interaction, particularly how AI systems can respond to evolving cognitive and emotional states to support trust formation and calibration in high-stakes environments.

\subsubsection*{Key Elements of Swift Trust}
 
To model and support swift trust effectively, it is useful to decompose it into key elements, as outlined in Table \ref{tab:trust_elements} and commonly cited in the literature 
\cite{Hoff2015, Cho2015, Endsley2023}. Among these, adaptability plays a critical role in high-stakes settings where user state and task demand shift rapidly. It reflects the AI system’s responsiveness to physiological, cognitive, and contextual signals, allowing it to adjust its explanations (e.g., simplifying content during stress) and behaviors (e.g., increasing feedback frequency during uncertainty) to maintain trust. As Cho et al. \cite{Cho2015} and Seong and Bisantz \cite{Seong2008} note, adaptive systems promote more accurate and timely trust calibration, allowing the user to rapidly align trust with the actual performance of AI. 

\begin{table}[h]
\centering
\caption{Key elements influencing trust in AI systems}
\label{tab:trust_elements}
\begin{tabular}{ll}
\toprule
\textbf{Element} & \textbf{Definition} \\
\midrule
Reliability & Perceived consistency and dependability of the AI actions \\
Competence & Perceived skill or ability of the AI to complete its task \\
Predictability & Operator’s ability to anticipate AI behavior based on context \\
Transparency & Clarity in how and why the AI makes decisions \\
Adaptability & The ability of AI to adjust its behavior and output based on evolving user states and task\\ & demands \\
\bottomrule
\end{tabular}
\end{table}

These elements are not static; rather, they are dynamically influenced by both user state and system behavior. An effective trust-supporting system must monitor changes in stress, workload, and emotional valence and adjust its communication strategy accordingly. Further, these elements can be modulated by explainability features, but only if the system is responsive to the underlying user state. For instance, low predictability can be improved by proactive explanations. Low transparency can be mitigated with “why” explanations about intent. Low reliability perception under stress may be best addressed with brief confidence statements (e.g., “High certainty: obstacle detected”) \cite{Endsley2023}. In the next section, we explore how explainability mechanisms can be leveraged to modulate these elements and support the dynamic formation and maintenance of swift trust. 

\subsection{Adaptive Explainability for Trust Formation }

Explainability has long been recognized as a mechanism for fostering trust in AI systems. However, most existing approaches are static and context agnostic, offering fixed explanations that do not adjust to the user’s state or task environment. Recent advances have begun to explore adaptive explanation strategies for instance, using reinforcement learning or partially observable Markov decision processes (POMDPs) to tailor explanations based on user type or task progression \cite{Sreedharan2018}. However, these approaches often rely on predefined user profiles or require explicit feedback, making them difficult to apply in high stakes, real time environments. Model reconciliation approaches \cite{Sreedharan2018} align AI explanations with human mental models, but typically assume static trust misalignment and do not account for fluctuating physiological or affective states. Floyd et al. \cite{Floyd2016} introduced trust guided transparency, where the AI modulates its behavior based on estimated user trust, but their system relied on explicit interaction logs and performance scores, rather than physiological signals. 

Recent research has demonstrated that trust related cognitive and emotional states can be inferred using physiological and behavioral signals such as heart rate variability (HRV), electrodermal activity (EDA), facial expressions, and gaze \cite{Khalid2016, rodriguez2023}. Fuzzy and neuro fuzzy models have been employed to classify trust states in real time, providing interpretable trust metrics for adaptive systems. However, these models have rarely been connected to explanation generation, leaving a gap in integrating trust estimation with communication behavior. 

Moreover, trust calibration aligning user trust with system capability is especially critical in high risk or time sensitive domains. Studies in aviation, medicine, and robotics have shown that miscalibrated trust leads to automation bias or disuse \cite{Hancock2011, Bobko2023}. Meanwhile, cognitive load plays a central role in explainability: too much detail can overwhelm the user, while too little can induce confusion or mistrust. Paleja et al. \cite{Paleja2021} found that tailoring explanation granularity benefits novice users under load but may frustrate experts, reinforcing the need for adaptive strategies that consider real time workload and user expertise. 

\subsection{User Centric and Affective XAI }
Early work on user aware XAI \cite{Soni2021} focused on clustering users by behavior patterns to personalize explanations. However, most approaches lack real time responsiveness, and few incorporate affective signals to dynamically adjust content. Ali et al. \cite{Ali2023} emphasize that explanations tailored to user context and emotional state enhance perceived competence and empathy, but current systems lack the infrastructure to connect affective inference with explanation logic. As summarized in Table \ref{tab:trust_adaptation_rules}, existing work highlights the individual importance of trust, explainability, and physiological modeling. However, few frameworks bring these together into a cohesive, real time trust adaptive explainability model that operates effectively in high stakes human AI teams.



\begin{table}[h]
\caption{Trust-relevant user states and corresponding explanation adaptation strategies}
\label{tab:trust_adaptation_rules}
\centering
\begin{tabular}{p{3.5cm} p{6cm} p{5.5cm}}
\toprule
\textbf{Trust-Relevant Factor} & \textbf{Empirical Insight} & \textbf{Explanation Adaptation Rule} \\
\midrule
High cognitive load & High workload reduces trust and task performance \cite{Hancock2011, Paleja2021} & Short, high-level summaries preferred. \\
High stress & Increased stress is negatively correlated with trust in automated systems\cite{Kohn2021_2} & Proactive, calming explanations delivered early. \\
Low emotional valence & Negative valence correlates with reduced trust \cite{Khalid2016} & Use reactive, empathetic tone or voice modality. \\
Low AI performance & Lowers reliability and competence perceptions \cite{Milivojevic2024, Hoff2015} & Provide corrective, fallback or reassuring explanations to recover trust. Reinforce competence with confidence indicators. \\
Low predictability & Hinders user ability to anticipate system behavior \cite{Milivojevic2024} & Provide rationale (why/how) behind actions. \\
Unfamiliar task (Expertise) & Increases mental effort; risks trust erosion \cite{Paleja2021} & Simplify content, emphasize goal relevance with proactive and guiding explanations. \\
Disengagement & Reduced engagement can impact performance and trust levels \cite{Hancock2011, Endsley2023, Paleja2021} & Switch to visual/interactive formats to regain attention. Mode of delivery can influence information absorption and processing cost. \\
\bottomrule
\end{tabular}
\end{table}

\subsubsection{Trust Modeling through Explainability Cues}
In Human Machine Teams (HMTs), trust and explainability are dynamically interlinked, unfolding as a sequence of cause-and-effect interactions. AI behaviors whether task related actions, feedback, or navigation decisions directly influence the human operator’s physiological state, emotional response, and cognitive processing. These changes, in turn, shape the operator’s perception of trust, impacting collaboration quality and task performance \cite{Shin2021}. These suggest how real-time adaptation of AI explanations, grounded in these causal pathways, can mitigate cognitive and emotional strain while reinforcing trust in high stakes and time sensitive environments. 

\textbf{AI Behavior and Trust Perception.} The observable behavior of AI such as decision making, task execution, or error handling shapes the operator’s perception of its competence, reliability, and intent. When the AI behaves transparently and contextually, users are more likely to perceive it as trustworthy. In contrast, opaque or inconsistent behaviors introduce uncertainty and distrust. Shin \cite{Shin2021} reports that causability and explainability account for over 58\% of variance in trust perceptions, underscoring the importance of clarity and communicative alignment in fostering trust. 

\textbf{Emotional and Physiological Responses.} Trust perceptions are further mediated by affective responses, which manifest physiologically through changes in heart rate (HR), heart rate variability (HRV), electrodermal activity, or neural activity (EEG) \cite{Mou2023}. For example, unexpected or ambiguous AI actions may trigger stress or frustration, while cooperative and predictable behavior fosters engagement and calm. These physiological markers serve as real time, implicit indicators of trust state \cite{Endsley2023}, enabling continuous user monitoring without explicit intervention. 

\textbf{Cognitive Load and Information Processing.} AI outputs that are complex, ambiguous, or mistimed can impose a high cognitive burden, impairing the user’s ability to process information and make timely decisions. This effect is especially pronounced in emergency response scenarios, where attention is divided, time is limited, and errors are costly. Prior studies \cite{Paleja2021, Endsley2023} show that cognitive overload negatively affects both trust and performance in collaborative human AI systems. Adaptive explainability can address this by modulating explanation complexity, timing, and content helping reduce overload, maintain attention, and recalibrate trust in real time. 
 
However, most XAI systems do not modulate their explanations based on implicit human signals such as stress, workload, or emotion, nor do they account for rapidly changing contextual cues. This limitation is particularly consequential in high-stakes environments, where cognitive overload and uncertainty diminish the operator’s ability to process static or overly generic explanations. There remains a significant gap in designing closed-loop adaptive XAI systems that leverage real-time physiological and contextual data to both interpret user states and dynamically tailor explanation strategies. Such systems would not only reflect the user’s cognitive and emotional state but also influence it over time supporting trust formation, cognitive efficiency, and collaborative resilience under pressure.

\section{Adaptive Explainability Trust Framework}
The findings presented in Section \ref{sec:relatedwork} lead to a critical insight: explainability is not merely a communication tool, but a dynamic mechanism for shaping and stabilizing swift trust. To support this, we propose, Adaptive Explainability Trust Framework (AXTF),  a conceptual framework designed to enhance human decision-making within human-AI teams by improving swift trust and team performance through dynamic adaptation of AI explanations based on real-time assessments of user states and environmental factors. Specifically, our framework enables AI explanations to be continuously tailored according to the user’s cognitive load, stress, and emotional state, as well as the task context, including urgency, goals, and environmental complexity, particularly in high-stakes domains such as emergency response.

\begin{figure}
  \centering
  \includegraphics[width=\linewidth]{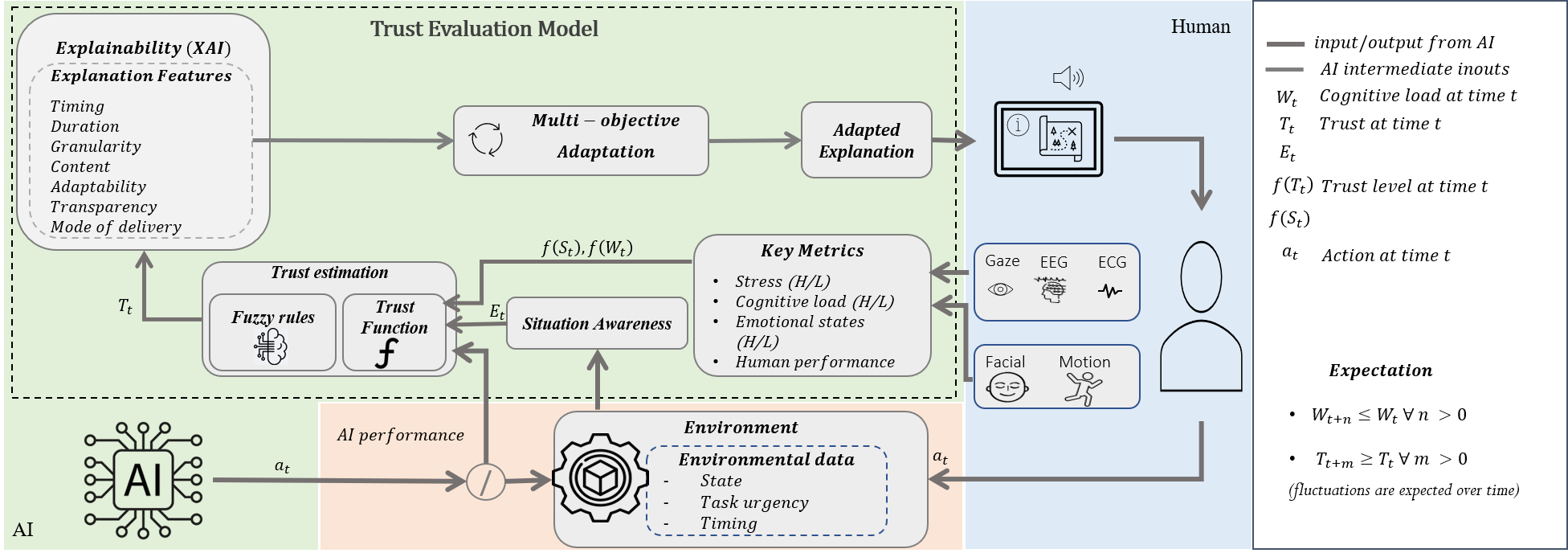}
  \caption{Adaptive Explainability Trust Framework (AXTF). The framework supports swift trust formation by dynamically adjusting explanation features—\textit{timing, duration, and granularity}—based on the user's cognitive load, emotions, and performance. This adaptive explainability reduces cognitive overload, enhances trust, and improves decision-making.}
  \label{fig:adaptive-model}
\end{figure}

To support swift trust and effective human AI teaming in high stakes environments, our proposed conceptual model links real time physiological and behavioral indicators of human state to adaptive explainability mechanisms. The model integrates three interconnected components: (1) multimodal feedback sensing and inference, (2) multi-objective trust modeling, and (3) explanation feature adaptation, with environmental inputs to ensure contextual relevance.

The proposed framework (Fig.~\ref{fig:adaptive-model}) forms a closed-loop pipeline that integrates real-time physiological and behavioral signals (e.g., EEG, ECG, heart rate variability—HRV) with environmental data such as task goals, urgency, and state to assess the user’s cognitive load (W), stress (S), and emotional valence (E) levels \cite{Paleja2021, Nishani2025}. Based on these user states and environmental performance metrics (e.g., task errors, success rates), dynamic trust estimation is performed using a multi-objective neurofuzzy rule-based inference engine. The framework then adapts key explanation features, including timing, duration, granularity, and mode of delivery, by mapping these trust estimates and contextual knowledge to reduce cognitive overload, enhance trust, and guide the subsequent human action ($a_{t+1}$). For example, in scenarios characterized by low trust and high cognitive load, short, moderate steps, and reactive explanations are more effective for fostering trust compared to lengthy, detailed ones. While some fluctuations in cognitive load ($W_t$) and trust ($T_t$) are expected, the ultimate goal of these adaptive explanations is to increase trust (T) and decrease cognitive load (W) over time, thereby promoting swift trust and improving situational awareness, decision-making, and overall team performance. This feedback-driven framework is designed to support temporal situational awareness, workload balancing, and trust resilience in high-pressure environments where explicit communication may be limited, but implicit signals provide actionable insight. The following subsections detail each component of the framework, including multimodal feedback sensing and inference, multi-objective trust modeling, and explanation feature adaptation.

\subsection{Real-Time Multimodal Inference of Human State}

The system continuously monitors a range of physiological and behavioral signals to infer latent cognitive and emotional states that are critical for trust assessment, including:

\begin{itemize}
    \item EEG and pupillometry for evaluating cognitive workload,
    \item ECG, galvanic skin response (GSR), and heart rate variability (HRV) to detect stress and arousal levels,
    \item Facial expressions, gaze tracking, and voice features to infer emotional valence and user engagement.
\end{itemize}

These raw signals undergo preprocessing and feature extraction pipelines before being classified into meaningful, interpretable states such as “High Workload” or “Low Valence” through specialized physiological inference models. By transforming complex biosignals into actionable, high-level indicators, the system enables downstream reasoning components to adapt interactions dynamically based on the user’s real-time cognitive and emotional state.



\subsection{Fuzzy Logic Based Trust Inference }
To translate these human state metrics into actionable trust estimates, we suggest a multiobjective neurofuzzy inference method. This approach allows the encoding of literature grounded, rule-based mappings (see Table 3) into an interpretable decision layer. 

Trust is estimated as a categorical variable with three levels Low, Medium, or High based on a combination of user physiological and affective states and system performance metrics. The model enables interpretable reasoning that links observed human state and AI behavior to well-established trust dimensions such as reliability, competence, predictability, transparency, and adaptability.  The model takes the following inputs:

\begin{itemize}
    \item \textbf{Input Variables}
    \begin{itemize}
        \item \textbf{Workload (W)}: categorized as Low, Medium, or High; derived from EEG features, gaze data, or behavioral task-switching patterns.
        \item \textbf{Stress level (S)}: categorized as Low, Medium, or High; inferred from ECG, GSR, and HRV metrics.
        \item \textbf{Emotion valence (E)}: categorized as Negative, Neutral, or Positive; estimated from facial expressions, tone of voice, or affective models.
        \item \textbf{System performance score (P)}: a normalized score between 0 and 1; computed from task metrics such as success rate and error frequency.
    \end{itemize}
    
    \item \textbf{Trust Output}
    \begin{itemize}
        \item \textbf{Trust (T)}: classified as Low, Medium, or High.
    \end{itemize}
\end{itemize}

\begin{table}[t]
\caption{Fuzzy rules for trust inference based on workload, stress, emotion, and performance}
\label{tab:fuzzy_rules}
\centering
\renewcommand{\arraystretch}{1.5}
\begin{tabularx}{\textwidth}{>{\raggedright\arraybackslash}p{5.5cm} X}
\textbf{Interpretation} & \textbf{Fuzzy Rule} \\
\toprule
Captures: low competence and predictability &
IF $W = \text{High}$ AND $S = \text{High}$ AND $E = \text{Negative}$ THEN $T = \text{Low}$ \\
Captures: reliability and transparency &
IF $S = \text{Low}$ AND $E = \text{Positive}$ AND $P > 0.8$ THEN $T = \text{High}$ \\
Captures: competence and predictability &
IF $W = \text{Low}$ AND $S = \text{Low}$ AND $E = \text{Neutral}$ THEN $T = \text{High}$ \\
Captures: adaptability &
IF $W = \text{Medium}$ AND $E = \text{Positive}$ AND $P > 0.6$ THEN $T = \text{Medium}$ \\
Captures: reliability under pressure &
IF $S = \text{High}$ AND $P < 0.4$ THEN $T = \text{Low}$ \\
Captures: transparency degradation due to overload/affect &
IF $W = \text{High}$ OR $E = \text{Negative}$ THEN $T = \text{Low}$ \\
Captures: balanced state supporting trust &
IF $W = \text{Medium}$ AND $S = \text{Medium}$ AND $E = \text{Positive}$ THEN $T = \text{High}$ \\
\bottomrule
\end{tabularx}
\end{table}








The fuzzy trust inference system maps normalized input variables into fuzzy linguistic categories using membership functions. Each input (e.g., workload, stress, emotion valence, performance) is associated with three fuzzy sets: \textit{Low}, \textit{Medium}, and \textit{High}, except for emotion valence, which uses \textit{Negative}, \textit{Neutral}, and \textit{Positive}.

\textbf{1. Workload (W), Stress (S)}

Both workload and stress are defined over the domain $[0, 1]$ and share the same triangular membership structure:

\[
\resizebox{\textwidth}{!}{$
\mu_{\text{Low}}(x) =
\begin{cases}
1, & x \leq 0.2 \\
\frac{0.5 - x}{0.3}, & 0.2 < x \leq 0.5 \\
0, & x > 0.5
\end{cases}
\qquad
\mu_{\text{Medium}}(x) =
\begin{cases}
0, & x \leq 0.2 \text{ or } x \geq 0.8 \\
\frac{x - 0.2}{0.3}, & 0.2 < x \leq 0.5 \\
\frac{0.8 - x}{0.3}, & 0.5 < x < 0.8
\end{cases}
\qquad
\mu_{\text{High}}(x) =
\begin{cases}
0, & x \leq 0.5 \\
\frac{x - 0.5}{0.3}, & 0.5 < x \leq 0.8 \\
1, & x > 0.8
\end{cases}
$}
\]

\textbf{2. Emotion Valence (E)}

Emotion valence is modeled over the range $[-1, 1]$:

\[
\resizebox{\textwidth}{!}{$
\mu_{\text{Negative}}(e) = 
\begin{cases}
1, & e \leq -0.5 \\
\frac{-e - 0.1}{0.4}, & -0.5 < e \leq -0.1 \\
0, & e > -0.1
\end{cases}
\quad
\mu_{\text{Neutral}}(e) =
\begin{cases}
0, & |e| > 0.5 \\
\frac{0.5 - |e|}{0.5}, & |e| \leq 0.5
\end{cases}
\quad
\mu_{\text{Positive}}(e) =
\begin{cases}
0, & e < 0.1 \\
\frac{e - 0.1}{0.4}, & 0.1 \leq e < 0.5 \\
1, & e \geq 0.5
\end{cases}
$}
\]

\textbf{3. System Performance (P)}

System performance is a normalized score $p \in [0, 1]$:

\[
\resizebox{\textwidth}{!}{$
\mu_{\text{Low}}(p) =
\begin{cases}
1, & p \leq 0.3 \\
\frac{0.5 - p}{0.2}, & 0.3 < p \leq 0.5 \\
0, & p > 0.5
\end{cases}
\quad
\mu_{\text{Medium}}(p) =
\begin{cases}
0, & p \leq 0.3 \text{ or } p \geq 0.7 \\
\frac{p - 0.3}{0.2}, & 0.3 < p \leq 0.5 \\
\frac{0.7 - p}{0.2}, & 0.5 < p < 0.7
\end{cases}
\quad
\mu_{\text{High}}(p) =
\begin{cases}
0, & p < 0.5 \\
\frac{p - 0.5}{0.3}, & 0.5 \leq p < 0.8 \\
1, & p \geq 0.8
\end{cases}
$}
\]

These functions are derived from literature indicating that increased workload and stress reduce trust in automation~\cite{Hancock2011, Endsley2023}, while positive valence and higher performance promote trust. Mapping continuous inputs into interpretable fuzzy categories supports transparent, adaptable trust modeling in real time.
 The fuzzy trust inference model estimates trust levels based on these real-time assessments of workload, stress, emotional valence, and system performance. Using a set of fuzzy rules (Table~\ref{tab:fuzzy_rules}) derived from empirical research and domain knowledge (see Table~\ref{tab:trust_elements}), the model captures complex interactions among these factors to produce a unified trust estimate \(T\) classified as Low, Medium, or High. This approach enables interpretable reasoning and supports adaptive AI behavior aligned with the operator’s current cognitive-affective state, improving human-AI collaboration.

\subsection{Trust Sensitive Explanation Adaptation } 

The model focuses on seven core explainability features that shape user experience and collaboration. We define these key features as follows: 
\begin{enumerate}
    \item \textbf{Timing:} Timely delivery of explanations is vital in high-pressure situations. Explanations can be delivered proactively, ahead of an AI action, to minimize potential confusion (e.g., "Avoiding unstable debris ahead"), or reactively, triggered by user hesitation or unexpected AI behavior. In an emergency context, timing must align with both the task phase and the user’s attentional bandwidth to avoid distraction or delay \cite{Paleja2021, Endsley2023, Shin2021}.

    \item \textbf{Duration:} The length of explanations should be carefully adapted to the time sensitivity of the situation and the user’s cognitive capacity. Under high cognitive load or time pressure,  brief explanations, typically lasting 2–3 seconds, are more effective in preserving situational focus and preventing distraction. On the other hand, when cognitive load is lower, longer or layered explanations can offer deeper insight without overwhelming the user \cite{Paleja2021, Endsley2023}. For instance, during search and rescue triage, the system may initially provide short verbal alerts, then follow up with optional elaboration once the situation stabilizes.

    \item \textbf{Granularity:} Explanation granularity refers to the level of detail provided in the explanation. High-level summaries, such as "Scanning lower level first," help reduce the user's information processing demands. In contrast, detailed step-by-step explanations, for example, "Entering sector B $\to$ mapping $\to$ thermal anomaly detected," are better suited for experienced users or situations where trust is high. The granularity should be adapted based on factors such as user familiarity, workload, and trust levels \cite{Paleja2021, Bobko2023, Zerilli2022} to ensure explanations remain cognitively accessible while still informative.

    \item \textbf{Content:} Explanation content is chosen based on task relevance and the user's current focus. Contextual or local explanations, such as "Rerouting due to obstacle," emphasize immediate actions or environmental conditions. In contrast, hierarchical explanations, for example, "Prioritizing lower floors due to heat signature density," communicate broader planning strategies. In emergency scenarios, providing contextual content enhances temporal situational awareness and responsiveness \cite{Setzu2021, Seong2008}.

    \item \textbf{Transparency:} Transparency shapes the user’s understanding of the AI's decision-making process. "How" transparency, such as "Based on heatmap and terrain risk, path updated," helps users evaluate the system’s methods. "Why" transparency, for example, "Avoiding risk to maximize coverage," clarifies the AI’s intent and goal reasoning. Both types support mental model alignment and trust; however, the appropriate level and form of transparency should be adapted based on the user's state and the context \cite{Chamola2023, Hoff2015, Floyd2016}.

    \item \textbf{Adaptability:} Adaptability acts as the central mechanism that dynamically modulates all other explainability features in real time. This capability allows AI systems to selectively tailor explanations to align with the user's current state and task objectives. Under conditions of high workload or stress, the system simplifies explanations and employs lower-detail modes to reduce cognitive load. Conversely, when users are calm and trust levels are high, the system can provide more complex and interactive explanations. This adaptability ensures that explanations remain both informative and sustainable, especially in high-pressure environments \cite{Lee2004, Hoff2015, Floyd2016, Soni2021}.

    \item \textbf{Mode of Delivery:} The medium used to deliver explanations significantly influences user comprehension and cognitive load. Visual formats such as maps, trajectory overlays, or alert icons are ideal when the user’s auditory attention is available. Textual explanations, including on-screen summaries or status updates, are more suitable during quieter moments or post-task phases. Auditory delivery through spoken instructions works best when the user’s hands and eyes are occupied. Combining multiple channels in a multimodal approach—such as spoken plus visual alerts—enhances system resilience and inclusivity. Research by Adadi and Berrada \cite{Adadi2018} demonstrates that multimodal explanations improve user understanding, particularly when users are multitasking or experiencing stress.

\end{enumerate}
\begin{table}[t]
\centering
\caption{XAI features aligned with trust and cognitive principles, and corresponding explanation strategies}
\label{tab:xai_features_strategies}
\begin{tabularx}{\textwidth}{p{3cm} p{4.2cm} X}
\toprule
\textbf{XAI Feature} & \textbf{Supports Swift Trust \& Cognitive Element(s)} & \textbf{Explanation Strategy Example} \\
\midrule
Timing & Predictability, Reliability & Provide proactive clarification during decision spikes or delays \\
Granularity & Competence, Cognitive Efficiency & Provide brief summary under high workload, detailed explanation when user trust is higher \\
Duration & Cognitive Efficiency, Predictability & Provide short explanations during time-critical phases, extended versions in low-pressure phases \\
Delivery Mode & Transparency, Engagement, Clarity & Switch between text, voice, or visual modes based on user attentiveness, emotional state, or context. Combine multiple modalities such as voice and visual to deliver context for enhanced information processing. \\
Content & Reliability, Context Sensitivity & Use local (contextual) explanations in dynamic settings, hierarchical in stable scenarios \\
Transparency & Transparency, Competence & Explain ``why'' decisions are made to improve model clarity and intention transparency \\
Adaptability & Adaptability, Personalization, Trust Calibration & Tailor content and modality dynamically based on inferred trust and workload signals \\
\bottomrule
\end{tabularx}

\end{table}


The inferred trust level is used to modulate these key explainability features. This trust adaptive mechanism supports continuous calibration, allowing the system to respond not just to performance, but to how the human feels and functions during collaboration. 

While explainability is often discussed as a means of improving user understanding or meeting regulatory requirements, in high stakes human AI teaming, it serves a deeper role: calibrating trust in real time. As trust is highly sensitive to changes in workload, stress, and affect, static or misaligned explanations may unintentionally erode confidence or overload the user. In contrast, adaptive explainability tuned to the user’s current cognitive and emotional state can serve as a powerful tool for trust repair, reinforcement, and regulation. Consider a search and rescue drone system operating in post-disaster environments. A human operator under high stress and workload may be overwhelmed by frequent decision updates. The system detects high HRV, elevated EEG load, and negative valence, infers low trust, and adapts by issuing short, confidence-framed explanations through audio (“Clear path detected. High certainty.”). If later states show reduced load and increased engagement, it shifts to detailed, interactive visualizations for planning and collaboration. 

\section{Future Directions}

While this work establishes a conceptual foundation for adaptive explainability in high-stakes human-AI teaming, several important avenues remain for future investigation. 

First, two-way communication between human and AI teammates should be more explicitly modeled not only to adapt AI explanations to user states, but also to enable reciprocal influence where the AI dynamically adjusts its behavior based on user reactions and evolving task demands. Such bi-directional interaction will allow the AI to both respond to human needs and evaluate and refine its own performance, closing the loop between perception, explanation, and behavior. Second, cultural and dispositional factors play a foundational role in trust formation but remain underexplored in adaptive XAI. Future research should investigate how traits such as uncertainty avoidance, communication preferences, and prior experience influence trust dynamics and explanation preferences, enabling more inclusive and culturally-aware explanation strategies. Third, implementation and evaluation in interactive, dynamic environments, such as simulation-based emergency response scenarios, will be essential to validate the framework. These settings offer controllable, high-fidelity contexts for assessing how real-time physiological and behavioral feedback impacts explanation effectiveness, trust calibration, and team performance under pressure. Fourth, advances in generative agent simulations \cite{Park2024} open opportunities for large-scale validation using synthetic populations embedded with memory, social reasoning, and behavioral diversity. These agent-based testbeds can be used to examine long-term trust trajectories and cross-profile adaptation strategies in simulated high-stakes team settings. 
Finally, to ensure AI safety in high-pressure decision environments, future work must incorporate safeguards that prevent explanation misuse or cognitive overload. Adaptive explanation systems must remain transparent, interpretable, and bounded by safety constraints that prevent miscalibration of trust—especially under uncertainty or stress. Embedding safety-aware logic into adaptation rules (e.g., thresholds on explanation complexity or delivery timing) will help maintain alignment with human cognitive capacity, trust boundaries, and ethical standards in mission-critical operations. 

In conclusion, advancing adaptive explainability through bi-directional interaction, cultural awareness, scalable simulation, and embedded safety principles will help realize the next generation of trust-sensitive, cognitively aligned, and ethically grounded human-AI systems.

\section{Conclusion and Contribution }
This work contributes to affective, situated, and trustworthy AI by introducing a conceptual framework for real-time adaptation of explainability to support swift trust and effective teamwork in high-stakes environments. The framework integrates multimodal implicit feedback physiological, behavioral, environmental, and contextual signals to infer user states such as workload, stress, and emotional valence. These inferred states inform the dynamic adjustment of explanation features (e.g., timing, granularity, modality), enabling alignment with the user’s cognitive and affective demands in pursuit of time-critical task goals. 
Explainability is reframed as a multi-objective adaptive function balancing transparency, cognitive efficiency, and trust calibration. The framework is designed to be model-agnostic and extensible, supporting the integration of diverse trust modeling and learning mechanisms. It enables AI systems to act as responsive teammates fostering trust, maintaining collaboration under pressure, and supporting decision-making when explicit communication is limited. This lays a foundation for generalizable real-world deployment of adaptive XAI in high-stakes domains such as emergency response, medical operations, and mission-critical decision support, where human-machine teaming must remain transparent, affect-aware, and cognitively efficient under uncertainty.

\section*{Declaration on Generative AI}

  
 During the preparation of this work, the author(s) used GPT-4 in order to: Grammar and spelling check. After using these tool(s)/service(s), the author(s) reviewed and edited the content as needed and take(s) full responsibility for the publication’s content. 

\newpage

\bibliography{main}




\end{document}